\documentclass[conference]{IEEEtran}

\IEEEoverridecommandlockouts

\usepackage{cite}
\usepackage{amsmath,amssymb,amsfonts}
\usepackage{algorithmic}
\usepackage{graphicx}
\usepackage{textcomp}
\usepackage{xcolor}
\usepackage{bm}
\usepackage{graphicx}
\usepackage{multirow}
\usepackage{multicol}
\usepackage{booktabs}
\usepackage{caption}
\usepackage{algorithm}
\usepackage{algorithmic}
\usepackage{subcaption}
\usepackage{float}
\usepackage{placeins}
\usepackage{comment}
\usepackage{xcolor}

\usepackage{geometry}

\newcommand{\norm}[1]{\left\lVert#1\right\rVert}

\def\BibTeX{{\rm B\kern-.05em{\sc i\kern-.025em b}\kern-.08em
    T\kern-.1667em\lower.7ex\hbox{E}\kern-.125emX}}

\begin{document}

\title{Mixture-of-Experts for Distributed Edge Computing with Channel-Aware Gating Function}

\author{
\IEEEauthorblockN{Qiuchen Song$^1$, Shusen Jing$^2$, Shuai Zhang$^3$, Songyang Zhang$^4$, and Chuan Huang$^1$}

\thanks{
This work was supported in part by the Nature Science Foundation of China under Grant 62341112, in part by the Key Project of Shenzhen under Grant JCYJ20220818103006013.}
\thanks{
Qiuchen Song is with the Shenzhen Future Network of Intelligence Institute, the School of Science and Engineering, and the Guangdong Provincial Key Laboratory of Future Networks of Intelligence, The Chinese University of Hong Kong, Shenzhen 518172, China (e-mail: qiuchensong@link.cuhk.edu.cn).}
\thanks{Shusen Jing is with Department of Radiation Oncology, University of California, San Francisco, CA, 94118, USA (e-mail: shusen.jing@ucsf.edu).}
\thanks{Shuai Zhang is with Department of Data Science, New Jersey Institute of Technology, Newark, NJ, 07102, USA (e-mail: sz457@njit.edu).}
\thanks{Songyang Zhang is with Department of Electrical and Computer Engineering, University of Louisiana at Lafayette, Lafayette, LA, 70504, USA (email: songyang.zhang@louisiana.edu).}
\thanks{
Chuan Huang is with the School of Science and Engineering, the Shenzhen Future Network of Intelligence Institute, and the Guangdong Provincial Key Laboratory of Future Networks of Intelligence, The Chinese University of Hong Kong, Shenzhen 518172, China (e-mail: huangchuan@cuhk.edu.cn).
}
}

\maketitle

\begin{abstract}
 In a distributed mixture-of-experts (MoE) system, a server collaborates with multiple specialized expert clients to perform inference. The server extracts features from input data and dynamically selects experts based on their areas of specialization to produce the final output. Although MoE models are widely valued for their flexibility and performance benefits, adapting distributed MoEs to operate effectively in wireless networks has remained unexplored. In this work, we introduce a novel channel-aware gating function for wireless distributed MoE, which incorporates channel conditions into the MoE gating mechanism. To train the channel-aware gating, we simulate various signal-to-noise ratios (SNRs) for each expert’s communication channel and add noise to the features distributed to the experts based on these SNRs. The gating function then utilizes both features and SNRs to optimize expert selection. Unlike conventional MoE models which solely consider the alignment of features with the specializations of experts, our approach additionally considers the impact of channel conditions on expert performance. Experimental results demonstrate that the proposed channel-aware gating scheme outperforms traditional MoE models.
\end{abstract} 

\begin{IEEEkeywords}
Mixture-of-experts, channel state information, learning for communications.
\end{IEEEkeywords}

\section{Introduction}
Next-generation wireless communications, such as sixth-generation (6G) cellular networks, are expected to enable massive, heterogeneous data processing tasks through artificial intelligence (AI). These advancements have been supporting diverse applications, including multimedia data processing, smart healthcare, and integrated sensing and communications (ISAC) \cite{tian2024performance}. Modern wireless systems consist of numerous edge devices equipped with local computational resources, facilitating the deployment of deep neural network models. Under the constraints of communication and computation resources, how to efficiently assign data to suitable edge devices to optimize performance, computational efficiency, and communication overhead emerges as a critical challenge.

The Mixture-of-Experts (MoE) paradigm leverages multiple specialized neural networks, or “experts,” that collaborate to process input data samples \cite{jordan1994hierarchical}. Through a learned gating mechanism that routes inputs to the most suitable experts, MoE models achieve high model capacity and computational efficiency \cite{shazeer2017outrageously}. This approach has demonstrated notable success in applications such as large language models (LLMs) and computer vision \cite{fedus2021switch}. Given its inherently distributed design, MoE is a promising approach for allocating computational resources across multiple devices in distributed wireless communication systems \cite{park2019wireless}. For example, a key application is Unmanned Aerial Vehicle (UAV)-assisted integrated sensing and communications (ISAC), where individual UAVs act as expert nodes. In this setting, UAVs collaboratively handle complex sensing and decision-making tasks while maintaining formation and coordinating missions \cite{wu2021enabled}. Another example is the wireless industrial Internet of Things (IIoT), where distributed sensors and processing units can form an expert network to support tasks like real-time manufacturing monitoring, predictive maintenance, and quality control \cite{liang2018industrial}.

As an emerging AI technique, MoE has seen limited exploration in wireless communication systems. In \cite{du2024mixture}, MoE is enhanced with LLMs to select appropriate experts of deep reinforcement learning models to make decisions in intelligent networks. Similarly, \cite{xue2024wdMoE} introduces an expert selection policy for LLMs in wireless settings, accounting for both performance and end-to-end latency. However, existing approaches often assume ideal channel conditions or treat expert selection and communication as separate problems, resulting in suboptimal performance in real-world wireless deployments \cite{yang2019federated}.
Other works adopt neural networks with MoE architecture for baseband signal processing \cite{van2024mean}\cite{gao2023moe} instead of addressing the communication requirements of the experts in the distributed MoE setting. Thus, how to enhance the efficiency of MoE considering real channel conditions remains a challenge.

To fully harness the potential of MoE in wireless systems, several challenges need to be addressed. First, the dynamic nature of wireless channels introduces uncertainty in expert accessibility and communication quality. Traditional MoE architectures, designed for stable network environments, may make inefficient routing decisions when experts experience variable channel conditions due to, for example, fading, interference, and mobility \cite{wang2020edge}. 
Additionally, the diverse computational capacities and energy constraints across wireless devices heighten the need for load balancing to mitigate bottlenecks and ensure optimal resource utilization \cite{zhou2019edge}. Limited bandwidth and latency constraints in wireless networks further necessitate efficient expert selection strategies that balance computational and communication costs \cite{letaief2019roadmap}.

To address the aforementioned challenges, in this work, we propose a novel channel-aware MoE architecture that incorporates channel state information for expert selection. Specifically, unlike traditional MoE systems that rely solely on feature-expert alignment, our gating network takes both data features and channel quality as inputs, enabling intelligent expert selection based on dynamic channel conditions and real-time computation capacity.

\begin{figure}[t]
    \centering
\includegraphics[width=0.4\textwidth]{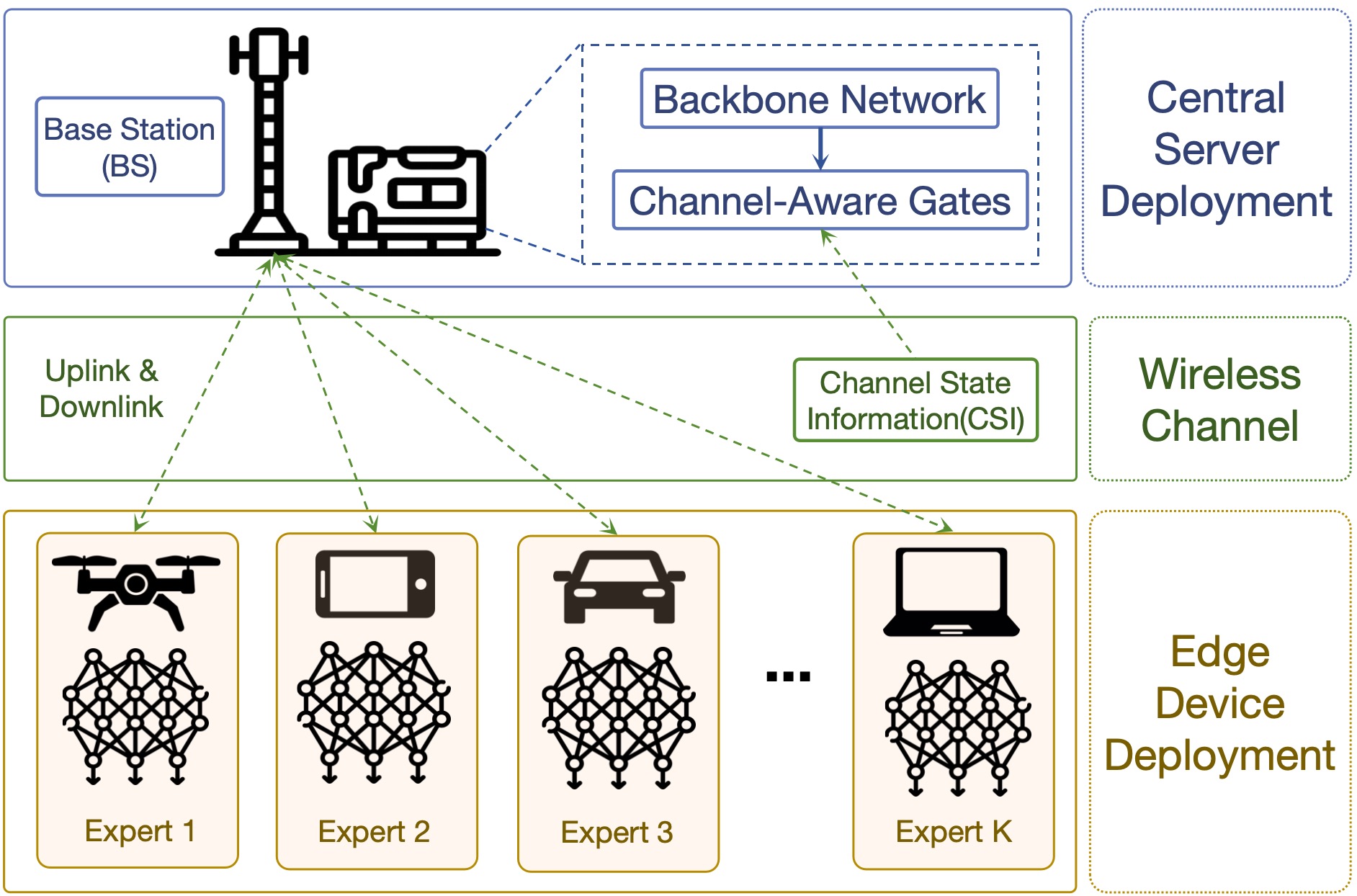}
    \caption{Structure of MoE-based Edge Computing: The system is deployed across the server of the base station and edge device in the wireless communication environment. The backbone network and gating network are operated at BS, while the expert networks are distributed across the edge devices}
    \label{fig:model}
\end{figure}

Our contributions can be summarized as follows:
\begin{itemize}
    \item We propose a MoE-based distributed edge computing framework with channel-aware gating function for expert selection. To the best knowledge of the authors, this is the first work leveraging both feature and channel condition for expert selection in distributed MoE. Previous works either assume ideal channel conditions for distributed MoE for their applications \cite{du2024mixture} \cite{xue2024wdMoE}, or apply centralized MoE (a neural network architecture) for baseband signal processing in communication systems \cite{van2024mean}\cite{gao2023moe}.
    \item We introduce a channel-aware gating network to enhance expert selection efficiency. This network processes concatenated feature and channel state information vectors through a compact dense neural network, generating routing soft decisions on expert selection, accounting for both the specialty and channel condition of each expert.
    \item We conduct extensive experiments validating the effectiveness of the proposed framework across different machine learning architectures, including ResNet-18 and Vision Transformer (ViT). Our analysis provides comprehensive benchmarks for MoE performance in both analog and digital settings, simulating realistic scenarios with dynamic fading, interference, and mobility.
    \item We conduct an ablation study to assess the impact of key parameters in the channel-aware MoE, offering practical insights and guidelines for real-world system deployment.
\end{itemize}
Our experimental results demonstrate the robustness of our proposed framework in addressing the wireless channel variations, while maintaining competitive accuracy compared to ideal channel conditions. 

The rest of the paper is organized as follows. Section \ref{sec:sys} describes the systems and provides preliminaries, after which we introduce the proposed channel-aware gating function in Section \ref{sec:propose}. Following the experimental results in Section \ref{sec:exp}, Section \ref{sec:con} concludes this work and suggests several future directions.

\section{System Description} \label{sec:sys}
Before venturing into the structure of our proposed channel-aware MoE framework, we first present preliminaries and illustrate our system setup.

\begin{figure*}[t]
    \centering
\includegraphics[width=0.55\textwidth]{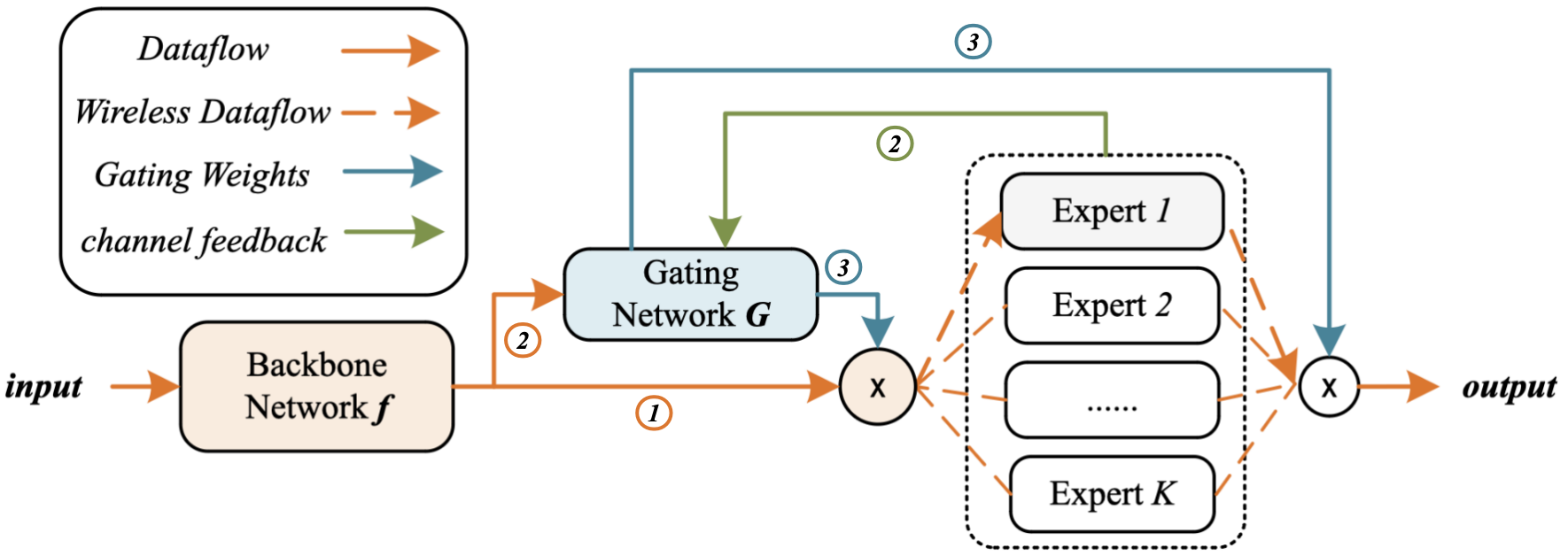}
    \caption{Data Workflow in the Channel-Aware MoE: In the channel-aware MoE, input data is initially processed by the backbone network into latent embeddings, which are then transmitted to the corresponding expert devices via wireless links, as directed by the gating network. The expert networks provide feedback on wireless channel conditions to the gating network, enhancing the robustness of both latent representation dispatching and expert output integration within the gating network.}
    \label{fig:workflow}
\end{figure*}

\subsection{MoE-based Distributed Edge Computing System}
In this work, we focus on a distributed edge computing system, where each edge device connects to the central server via a wireless link. Each edge device contains a computation unit, where deep neural network models can be deployed. Note that, considering the heterogeneity of local models and the unbalance of the computation capacities, the deep neural networks deployed in the edge device shall diverge from each other, which serves as the experts in the MoE system.
To fully utilize the computation resource at the edge end, the collected data is first processed through the backbone network at the base station (BS) and then distributed to an expert for data-oriented feature processing, after which the processed features are sent back to the BS as the final results. The overall system structure is depicted as Fig. \ref{fig:model}.

Next, we introduce the \textit{mathematic formulation} and \textit{notations} of our channel-aware MoE system.
Consider a distributed MoE system consisting of a server and $K$ experts. 
The server contains a backbone 
$F(\cdot;\theta_F): \mathbb{R}^{N_{in}} \rightarrow  \mathbb{R}^{N_{F}}$ 
network and a gating network 
$G(\cdot ;\theta_G): \mathbb{R}^{N_{F}} \rightarrow  \Delta_K$, 
which are parameterized by different neural networks (NNs) with weights $\theta_F$ and $\theta_G$, respectively. Here $N_{in}$, $N_F$ and $\Delta_K$ denote the data dimension, feature dimension and $K$-dimension simplex. Each expert possesses a specialized NN 
$Q_k(\cdot; \theta_{Q_k}): \mathbb{R}^{N_{F}}\rightarrow \mathbb{R}^{N_{out}}$ parameterized by weights $\theta_{Q_k}$, 
where $N_{out}$ denotes the output dimension of each expert.

During the inference stage, a data sample $x\in \mathbb{R}^{N_{in}}$ is first processed by the backbone $F(\cdot;\theta_F)$ to extract features, denoted by
\begin{gather}
z = F(x;\theta_F),
\end{gather}
where $z$ is the feature. 
Then, the gating function $G(\cdot ;\theta_G)$ chooses an expert to finalize the inference based on the alignment of the feature and the specialty of the experts, i.e., 
\begin{gather}
\Bar{k} = \arg\max G(z;\theta_G)    
\end{gather}
where $\Bar{k}$ is the index of the selected clients. With the selection of suitable experts, the feature $z$ is sent to the corresponding expert $\Bar{k}$ to perform specialized processing via
\begin{gather}
o_{\Bar{k}} = Q_{\Bar{k}}(z; \theta_{Q_{\Bar{k}}}).
\end{gather}

Finally, the processed features $o_{\Bar{k}}$ are transmitted back to the BS for feature aggregation and final task implementation.
Generally, the training loss of an MoE can be expressed as
\begin{gather}\label{eq:naive}
\begin{aligned}
\mathcal{L}(F, G, \{Q_k\}_{k=1}^K)= \mathsf{E}_{x,y,z=F(x)}l\left(Q(z\mathbf{1}^T)G(z),y\right),
\end{aligned}
\end{gather} 
where $Q(Z)=[Q_1(Z[1]),Q_2(Z[2]),\cdots, Q_K(Z[K])]\in \mathbb{R}^{ N_{out} \times K}$ is the concatenation of experts' outputs with the corresponding columns of $Z$ as input. Here, $Z[k]$ denotes the $k$-th column and $l(\cdot, \cdot)$ denotes the sample loss. $G(F(x))\in \mathbb{R}^K$ refers to the weight of each expert's decision, where $\mathbf{1}$ is the all-one vector.
The expectation $\mathsf{E}_{x,y,z=F(x)}$ accounts for drawing data points 
from the training dataset, which is averaged for the finite datasets. 
In this case, all of the experts take the same input $z$, and the notation $Q$ is defined for convenience in later use. 
Recall that $F$, $G$ and $\{Q_k\}_k$ are functions parameterized by NNs. Thus, the minimization of this loss essentially optimizes the corresponding weights $\theta_F$, $\theta_G$ and $\{\theta_{Q_k}\}_k^K$.
Notably, during inference, only one single expert is selected in our framework to compute the final results, whereas during training, the final output is obtained by calculating the weighted sum of all experts' outputs, which is then used to calculate the loss and update the weights.

\subsection{Wireless Channel Impacts}
In wireless scenarios, the transmitted feature $z$ shall be distorted by the wireless channel degrading the performance of MoE. Suppose that the $\Bar{k}$-th expert is selected. It shall receive the signal, denoted by
\begin{gather}\label{eq:dep1}
y_{\bar{k}} = h_{\bar{k}}pz + n_k
\end{gather}
where $h_{\bar{k}}\in \mathbb{R}$ accounts for the wireless fading, $p$ is the power scaling factor, and $n_{\bar{k}}$ is the noise vector,  whose entries are independently identically distributed (i.i.d.) \emph{Gaussian} with zero mean and $\sigma_k^2$ variance. The distorted feature, defined as
\begin{gather}\label{eq:dep2}
\tilde{z}_{\bar{k}} \triangleq y_{\bar{k}}/(ph_{\bar{k}})    
\end{gather}
should be used by the expert to compute the specialized results $Q_k(\tilde{z})$. 
Here are several additional assumptions in our MoE-based system. First, the server controls the average transmitted power, thus $p$ is a constant known by the experts. Second, the experts have perfect knowledge about the channel state information (CSI) $h_{\bar{k}}$. Third, the communication between the server and experts is analog. Nevertheless, our experiments include both results in analog and digital communication under the fading channels.

\section{Proposed Channel-Aware Gating Network}\label{sec:propose}
In practical scenarios, the channel condition directly affects the accuracy of the received feature and the final output from the selected expert. 
For example, an expert with the best alignment between feature and specialty but a very poor channel may perform worse than an expert with secondary alignment but a good channel. Therefore, in a wireless distributed MoE, both channel quality and feature-expert alignment shall be considered for model training.

For clarification, we assume the channel between the server and experts is ideal during the training, for example when using cable connections. During the inference, the channel is noisy, leading to distortion of the features.

\subsection{Channel-aware Gating}
Suppose that the server has the perfect knowledge of the downlink CSI $\{h_k\}_{k=1}^K$ and noise power $\{\sigma^2_k\}_{k=1}^K$. Then, the distortion variance can be calculated, according to Eq. (\ref{eq:dep1}), as 
    $\tilde{\sigma}_k^2 = \frac{\sigma^2}{p^2h_k^2}$.
The channel-aware gating is defined as $G_{AW}(\cdot, \cdot;\theta_{AW}): \mathbb{R}^{N_F}\times \mathbb{R}^{K}\rightarrow \Delta_K, (z, \tilde{\sigma})\mapsto G_{AW}(z, \tilde{\sigma}),$ 
where $\tilde{\sigma}$ is the concatenation of  $\{\tilde{\sigma}_k\}_{k=1}^K$. Given fixed (pretrained) $F$ and $\{Q_k\}_{k=1}^K$, the loss function for training $G_{AW}$ is calculated by
\begin{gather}\label{eq:awMoE}
\begin{aligned}
&\tilde{\mathcal{L}}(G_{AW}\!)\! = \! \mathsf{E}_{\tilde{\sigma}, n}\mathsf{E}_{y,z}l\!\left(Q(z\mathbf{1}^T\!\!\!+ n\text{diag}(\tilde{\sigma}))G_{AW}\!(z,\tilde{\sigma}),y\right),   
\end{aligned}  
\end{gather}
where $\tilde{\sigma} = [\tilde{\sigma}_1, \tilde{\sigma}_2,..., \tilde{\sigma}_K]\in \mathbb{R}^K$ and $n\in \mathbb{R}^{N_F\times K}$ is i.i.d. standard \emph{Gaussian} matrix. Ideally, minimizing this loss function leads to the optimal $G_{AW}$ independent to the distribution of $\tilde{\sigma}_k$, thus $h_k$ and $\sigma^2$. To obtain the ideal $G_{AW}$, we first derive the following lower bound
\begin{gather}
\begin{aligned}
&\min_{G_{AW}}\tilde{\mathcal{L}}(G_{AW}) \\
&\geq \mathsf{E}_{z}\mathsf{E}_{\tilde{\sigma}}\min_{G_{AW}}\mathsf{E}_{y|z}\mathsf{E}_nl\left(Q(z\mathbf{1}^T\!\!\!+ n\text{diag}(\tilde{\sigma}))G_{AW}(z,\tilde{\sigma}),y\right) \\
&\geq \mathsf{E}_{z}\mathsf{E}_{\tilde{\sigma}}\min_{\pi\in\Delta_K}\mathsf{E}_{y|z}\mathsf{E}_nl\left(Q(z\mathbf{1}^T\!\!\!+ n\text{diag}(\tilde{\sigma}))\pi,y\right),
\end{aligned}
\end{gather}
where the first inequality results from the swap of minimum and expectation. This further indicates the ideal $G_{AW}$ can be optimized via
\begin{gather}
G_{AW}^*(z,\tilde{\sigma}) = \arg\min_{\pi\in\Delta_K} \mathsf{E}_{y|z}\mathsf{E}_{n}l\left(Q(z\mathbf{1}^T\!\!\!+ n\text{diag}(\tilde{\sigma}))\pi,y\right),\nonumber
\end{gather}
which does not depend on the distribution of $\tilde{\sigma}$. The independence of channel statistics highly benefits the generalization of our model in real scenarios.


\subsection{Two-stage Training of Channel-aware MoE}
In the first stage, we train the entire MoE system with the assumptions of perfect channels using the training loss in Eq. (\ref{eq:naive}). The obtained weights, i.e., $\theta_F, \theta_G$, and  $\{\theta_{Q_k}\}_{k=1}^K$ serve as the initialization for the second stage. During the training, the gating output tends to collapse to a single expert despite the random initialization, which degrades the performance of MoE. This phenomenon has been observed in many existing works \cite{zhou2022mixture,chi2022representation}. To prevent the collapse, we adopt a balanced regularization similar to \cite{fedus2021switch} during the training, denoted by
\begin{gather}\label{eq:blance}
\mathcal{R}_{bal}(\theta_G) = \norm{\frac{1}{B}\sum_{i=1}^BG(x^i;\theta_G)-\frac{1}{K}}^2,
\end{gather}
where $B$ denotes the batch size and $x_i$ denotes the $i$-th sample in the batch. 

In the second stage, we introduce simulated wireless channels to replace the naive gating network with a channel-aware network while keeping the backbone and expert network parameters frozen. Particularly, we randomly simulate a set of sufficient channel conditions and train the MoE using the loss function presented in Eq. (\ref{eq:awMoE}). 

The detailed training process of the proposed channel-aware MoE is illustrated as Algorithm \ref{alg:training}.

\begin{algorithm}[t]
\caption{Two-Stage Training of Channel-Aware MoE}
\label{alg:training}
\begin{algorithmic}[1]
\REQUIRE Training dataset $\mathcal{D}$, backbone $F(\cdot;\theta_F)$, channel-aware gating $G_{AW}(\cdot,\cdot; \theta_{AW})$ and K experts $\{Q_k(\cdot;\theta_{Q_k})\}_{k}^{K=1}$
\ENSURE $\theta_F$, $\theta_{AW}$ and $\{\theta_{Q_k}\}_{k=1}^K$
\STATE // Stage 1: Pre-training $F$ and $\{Q_k\}_{k}^{K=1}$. 
\FOR{each epoch}
    \FOR{samples the $i$-th sample $(x^i, y^i)$ in batch}
        \STATE $z^i \leftarrow F(x^i;\theta_F)$
        \STATE $g^i \leftarrow G(z^i;\theta_G)$ 
        \STATE $\hat{y}^i \leftarrow \sum_{k=1}^K g^i[k]Q_k(z^i)$
    \ENDFOR
    \STATE Update $\theta_F$, $\theta_{G}$ and $\{\theta_{Q_k}\}_{k=1}^K$ through SGD over batch loss $\mathcal{L}_{B} = \frac{1}{B}\sum_{i=1}^Bl(\hat{y}^i, y^i)+\lambda\mathcal{R}_{bal}$, where $\mathcal{R}_{bal}$ is defined in eq. (\ref{eq:blance}).
\ENDFOR

\STATE // Stage 2: Train a channel-aware gating $G_{AW}$ to replace $G$, with backbone $F$ and experts $\{Q_k\}_{k=1}^K$ frozen.
\FOR{each epoch}
    \FOR{samples the $i$-th sample $(x^i, y^i)$ in batch}
        \STATE $z^i \leftarrow F(x^i;\theta_F)$
        \FOR{each expert $k = 1$ to $K$}
            \STATE Randomly generate $\tilde{\sigma}^i_k$ or, instead, generate $(h^i_k, \sigma^i_k)$ and calculate  $\tilde{\sigma}^i_k = \frac{\sigma_k^i}{p|h^i_k|}$.
            \STATE $\tilde{z}^i_k = z^i + \tilde{\sigma}^i_kn^i_k$, where $n_k$ is standard \emph{Gaussian}
        \ENDFOR
        \STATE $g^i \leftarrow G_{AW}(z^i, \{\tilde{\sigma}^i_k\})$ \COMMENT{Channel-aware gating}
        \STATE $\hat{y}^i \leftarrow \sum_{k=1}^K g^i[k] Q_k(\tilde{z}^i_k)$
    \ENDFOR
    \STATE Update $\theta_{AW}$ through SGD over batch loss $\mathcal{L}_{B} = \frac{1}{B}\sum_{i=1}^Bl(\hat{y}^i, y^i)+\lambda\mathcal{R}_{bal}$.
\ENDFOR
\end{algorithmic}
\end{algorithm}

\section{Experimental Results}\label{sec:exp}
In this section, we present the experimental results of the proposed channel-aware MoE, comparing to the conventional MoE
with naive gating, under different channel conditions and wireless setups.

\begin{table}[t]
\centering
\caption{Network Architecture Comparison}
\begin{tabular}{cccc}
\toprule
Dataset & Backbone & Gating & Expert \\
\midrule
\multirow{2}{*}{CIFAR-10}
& LeNet-5 & 120-64-$K$ MLP & 120-84-10 MLP \\
\cmidrule(lr){2-4}
& ResNet-18 & 512-256-$K$ MLP & 512-10 MLP \\
\midrule
CIFAR-100 
& ViT & 384-256-$K$ MLP & 384-1536-384 MLP \\
\bottomrule
\end{tabular}
\end{table}

\subsection{Experimental Setup}
In this work, we test the MoE structure based on several common NN architectures, varying the number of experts from $K=4$ to $K=12$ in different datasets. During the training of the channel-aware gating network, we consider a Rayleigh fading channel for $\tilde{h}_k$, and we sample the SNRs (in dB) for $\tilde{z}_k$ from a \emph{Gaussian} distribution $\sim \mathcal{N}(30, 2,500)$. 
To make the channel-ware gating encounter a variant of combinations of $\{\tilde{\sigma}_k\}_k$, we utilize a large range in the distribution of SNR selection,
which could benefit the generalization of the trained NN when serving in different wireless networks. All the experiments are conducted on NVIDIA A100 GPUs. 

\subsubsection{Models and Datasets}
We evaluate the performance of the proposed channel-aware MoE training schemes, with backbone networks designed as
Lenet-5 \cite{lecun1998gradient}, ResNet-18 \cite{he2016deep} and Vision Transformer (ViT) \cite{dosovitskiy2020image} for image classification. More specifically, we evaluate the performance of Lenet-5 and ResNet-18 backbones in CIFAR10, and ViT-based MoE in CIFAR100 \cite{krizhevsky2009learning}. In the experiment, we split the dataset as {75\%/8.3\%/16.7\%} for training/validation/testing.
\begin{figure}[t]
    \centering
\includegraphics[width=0.65\columnwidth]{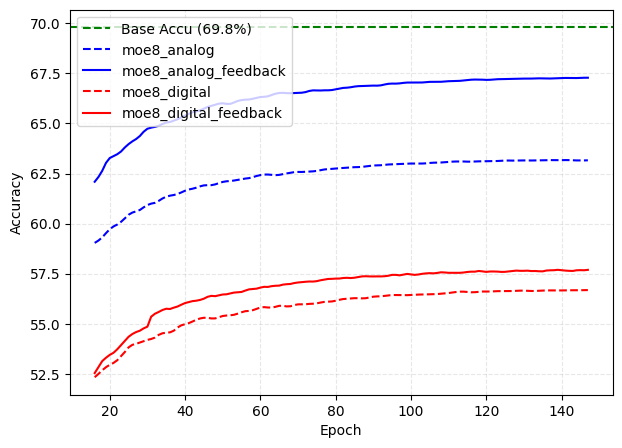}
    \caption{Classification accuracy of Lenet-5 on CIFAR-10 dataset}
    \label{fig:lenet}
\end{figure}
\begin{figure}[t]
    \centering
  \includegraphics[width=0.65\columnwidth]{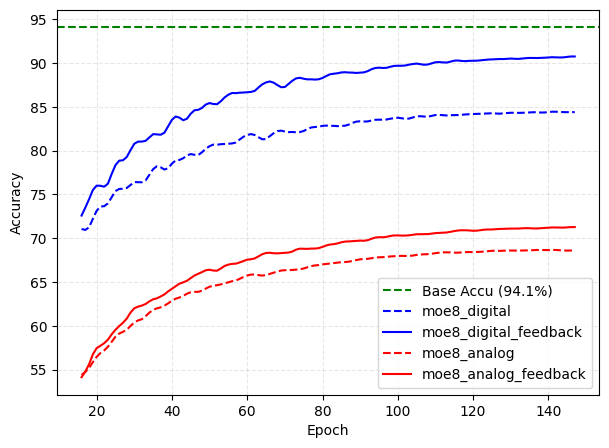}
    \caption{Classification accuracy of ResNet-18 on CIFAR-10 dataset}
    \label{fig:resnet}
\end{figure}
\begin{figure}[t]
    \centering
    \includegraphics[width=0.65\columnwidth]{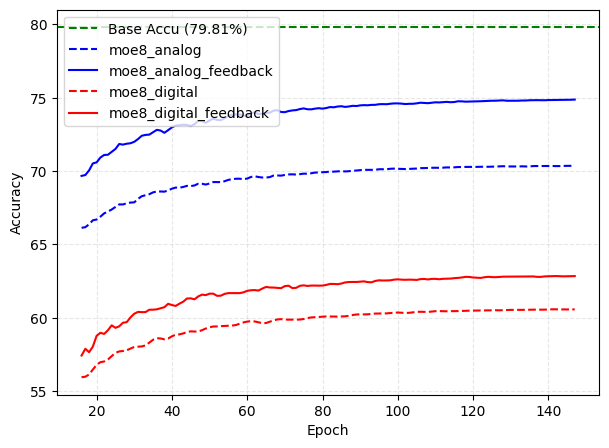}
    \caption{Classification accuracy of ViT on CIFAR-100 dataset}
    \label{fig:vit}
\end{figure}

\begin{table*}[t]
\centering
\caption{Classification accuracy comparison under different communication scenarios with $K = 8$ experts.}
\begin{tabular}{llcccc}
\toprule
Scenario & Model & Dataset & Base Acc. & Naive Gating & Channel-Aware Gating\\
\midrule
\multirow{3}{*}{Analog} 
& Lenet-5 & CIFAR-10 & 69.8\% & 62.8\% & 67.5\% \\
& ResNet-18 & CIFAR-10 & 94.1\% & 84.7\% & 91.1\% \\
& ViT & CIFAR-100 & 79.81\% & 70.1\% & 74.9\% \\
\midrule
\multirow{3}{*}{Digital} 
& Lenet-5 & CIFAR-10 & 69.8\% & 56.6\% & 57.7\% \\
& ResNet-18 & CIFAR-10 & 94.1\% & 68.3\% & 71.0\% \\
& ViT & CIFAR-100 & 79.81\% & 58.2\% & 62.0\% \\
\bottomrule
\end{tabular}
\label{table2}
\end{table*}

\subsubsection{Analog and Digital Communication}
In this work, we test the proposed frameworks in both analog and digital communication systems.
In the digital communication simulation for latent vectors transmission, the input latent representation $z$ is normalized and quantized to 8-bit precision according to
\begin{equation}
    z_{quant} = \text{round}\left(\frac{(z - z_{min})(2^{8}-1)}{z_{max} - z_{min}}\right).
\end{equation}

Then, the quantized data undergoes the Huffman coding for source compression, followed by convolutional code with rate $R = \frac{1}{2}$ for error protection. The coded latent is modulated using the 16-QAM scheme. At the receiver, after demodulation and channel decoding, the latent vector is reconstructed through
\begin{equation}
    z_{recon} = \frac{z_{quant}(z_{max} - z_{min})}{2^{8}-1} + z_{min}.
\end{equation}





\subsection{Performance Evaluation}
In the test, we utilize the performance of MoEs under perfect channel conditions as the baseline, which can be obtained after training stage 1. Table \ref{table2} presents the results of classification accuracy of MoEs with naive and our channel-aware gating in both analog and digital communications. Moreover, we showcase the performance of the MoE models after different training epochs in the training stage 2 in Figs. \ref{fig:lenet}-\ref{fig:vit}. 
Note that the training phase does not involve digital communications, which is only implemented during the test phase after each epoch of training. From the results, we have the following observations. The imperfect channels degrade MoE performance under both analog and digital communication schemes compared with baseline, especially with the naive gating without knowledge of channel condition. 
The proposed channel-aware training scheme improves the performance of MoE under both analog and digital communications, compared to naive gating. Although the training phase does not involve digital communication, $\tilde{\sigma}_k$ can still indicate the feature distortion after digital transmission, from which channel-aware gating can make comprehensive decisions with the consideration of both feature-specialty alignment and feature error.
Interestingly, the accuracy of MoE with naive gating also increases during the training stage 2 when the backbone and experts are fixed. This may result from the situation that, the gating network favors experts with better robustness to the feature distortions.
In our experimental results, the MoEs perform worse in digital communications compared to the analog ones, which might be caused by weak channel coding and harsh channel conditions.


\subsection{Ablation Study}
\begin{figure}[t]
    \centering
\includegraphics[width=0.42\textwidth]{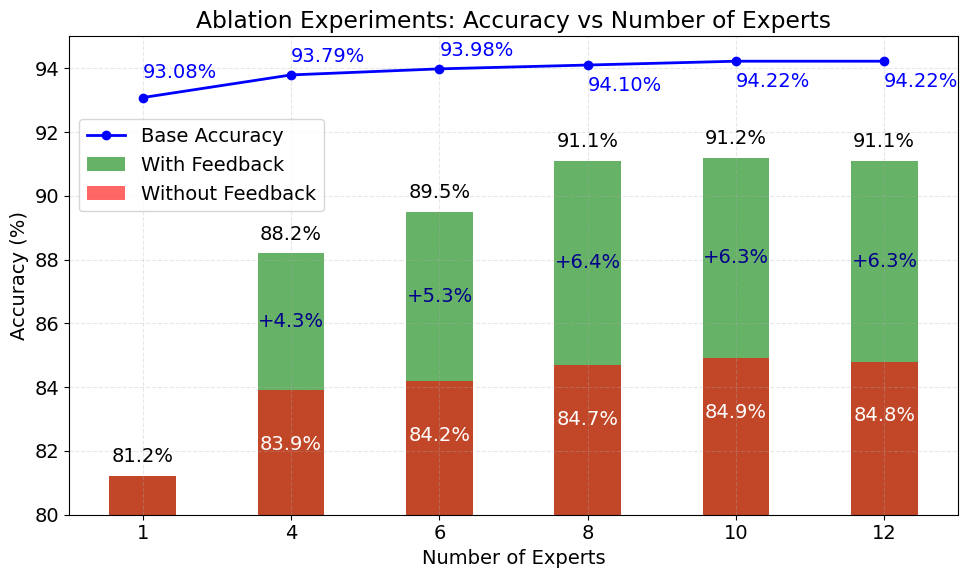}
    \caption{Accuracy comparison of different numbers of experts: With the increase of expert number, the improvement from channel-aware gating also rises, which reaches a certain bound with enough experts.}
    \label{fig:accuracy-comparison}
\end{figure}

We also conduct the ablation study on MoE with a different number of experts, with results showing in
Fig. \ref{fig:accuracy-comparison}.
This experiment adopts the ResNet-18 in Cifar-10 data. From the results, the increase of the expert number could lead to an improvement in the MoE performances, which provides sufficient specialization for different distributions. However, the performance saturates after having sufficient experts. As shown in Fig. \ref{fig:accuracy-comparison}, the performance improvement of MoE under imperfect channels exceeds that of the baseline. This can be attributed to the increased number of experts, which provides more options with favorable channel conditions. However, this also leads to greater overlap in the specializations among experts. For instance, adding more experts typically involves introducing ones with similar specialties to those already present, which does not necessarily enhance baseline performance. Nevertheless, duplicated experts allow the gating mechanism to select the one with the best channel condition at any given time, thereby improving overall performance.

\section{Conclusion}\label{sec:con}
In this work, we introduce a novel wireless edge computing system based on Mixture-of-Experts architecture, where a channel-aware gating function is introduced for model training considering both feature alignment and channel conditions. To provide a guideline for applying MoE in practical wireless scenarios, we experiment in several well-known deep learning architectures, including Lenet-5, ResNet and ViT, for image classification in both analog and digital communication systems, where a comprehensive ablation study is offered for system analysis. Our experimental results demonstrate the power of the proposed channel-aware gating function to capture the channel condition, as well as the efficiency of the proposed MoE-based wireless system. 
In our future works, we plan to explore tokenization techniques for representation embedding. Furthermore, although our current experiments focus on image classification tasks, the proposed channel-aware gating mechanism is task-agnostic and can also be extended to other scenarios such as natural language processing or real-time decision-making in autonomous system such as self-driving vehicle or UAV via deep reinforcement learning.


\bibliographystyle{IEEEtran}
\bibliography{main}

\end{document}